\documentclass[conference]{IEEEtran}
\IEEEoverridecommandlockouts
\usepackage[colorlinks=true, citecolor=red]{hyperref}
\usepackage{cite}
\usepackage{amsmath,amssymb,amsfonts}
\usepackage{algorithmic}
\usepackage{caption}
\usepackage{graphicx}
\usepackage{lscape}
\usepackage{multirow}
\usepackage{graphicx}
\usepackage{subcaption}
\usepackage{booktabs}
\usepackage{textcomp}
\usepackage{xcolor}
\def\BibTeX{{\rm B\kern-.05em{\sc i\kern-.025em b}\kern-.08em
    T\kern-.1667em\lower.7ex\hbox{E}\kern-.125emX}}
\begin{document}

\title{FOD-S2R: A FOD Dataset for Sim2Real Transfer Learning based Object Detection}

\author{
\IEEEauthorblockN{
Ashish Vashist\IEEEauthorrefmark{1}\hspace{1.5em}
Qiranul Saadiyean\IEEEauthorrefmark{1}\hspace{1.5em}
Suresh Sundaram\IEEEauthorrefmark{1}\hspace{1.5em}
Chandra Sekhar Seelamantula\IEEEauthorrefmark{2}
}
\IEEEauthorblockA{\IEEEauthorrefmark{1}Department of Aerospace Engineering, Indian Institute of Science (IISc), Bangalore, India}
\IEEEauthorblockA{\IEEEauthorrefmark{2}Department of Electrical Engineering, Indian Institute of Science (IISc), Bangalore, India}
}

\maketitle

\begin{abstract}
Foreign Object Debris (FOD) within aircraft fuel tanks presents critical safety hazards including fuel contamination, system malfunctions, and increased maintenance costs. Despite the severity of these risks, there is a notable lack of dedicated datasets for the complex, enclosed environments found inside fuel tanks. To bridge this gap, we present a novel dataset, FOD-S2R, composed of real and synthetic images of the FOD within a simulated aircraft fuel tank. Unlike existing datasets that focus on external or open-air environments, our dataset is the first to systematically evaluate the effectiveness of synthetic data in enhancing the real-world FOD detection performance in confined, closed structures. The real-world subset consists of 3,114 high-resolution HD images captured in a controlled fuel tank replica, while the synthetic subset includes 3,137 images generated using Unreal Engine. 
The dataset is composed of various Field of views (FOV), object distances, lighting conditions, color, and object size. Prior research has demonstrated that synthetic data can reduce reliance on extensive real-world annotations and improve the generalizability of vision models. Thus, we benchmark several state-of-the-art object detection models and demonstrate that introducing synthetic data improves the detection accuracy and generalization to real-world conditions. These experiments demonstrate the effectiveness of synthetic data in enhancing the model performance and narrowing the Sim2Real gap, providing a valuable foundation for developing automated FOD detection systems for aviation maintenance.
\end{abstract}

% \begin{IEEEkeywords}
% component, formatting, style, styling, insert
% \end{IEEEkeywords}

\section{Introduction}
Recent advances in aviation have significantly improved flight safety, operational efficiency, and airport infrastructure. However, as air traffic has increased, safety concerns related to airport environments have become increasingly critical. One of the persistent threats to aviation safety is Foreign Object Debris (FOD), which includes any loose object such as nuts, bolts, tools, or metal fragments that can damage aircraft during takeoff, landing, or ground operations. Small debris such as metal shavings can have severe consequences, including engine failure, fuel leaks, and system malfunctions, ultimately compromising passenger safety and flight integrity.

While modern computer vision and AI-based object detection techniques have significantly advanced the ability to identify foreign objects in open environments, such as runways and airport aprons, a major challenge remains in detecting FOD in constrained and enclosed areas, such as inside fuel tanks or maintenance bays \cite{farooq2024improved}\cite{munyer2021fod}\cite{zhao2019objectdetectiondeeplearning}. These spaces often have poor lighting, complex surfaces, irregular geometries, and limited accessibility, making conventional detection methods less effective.

\begin{figure}[htbp]
    \centering
    \includegraphics[width=1\linewidth]{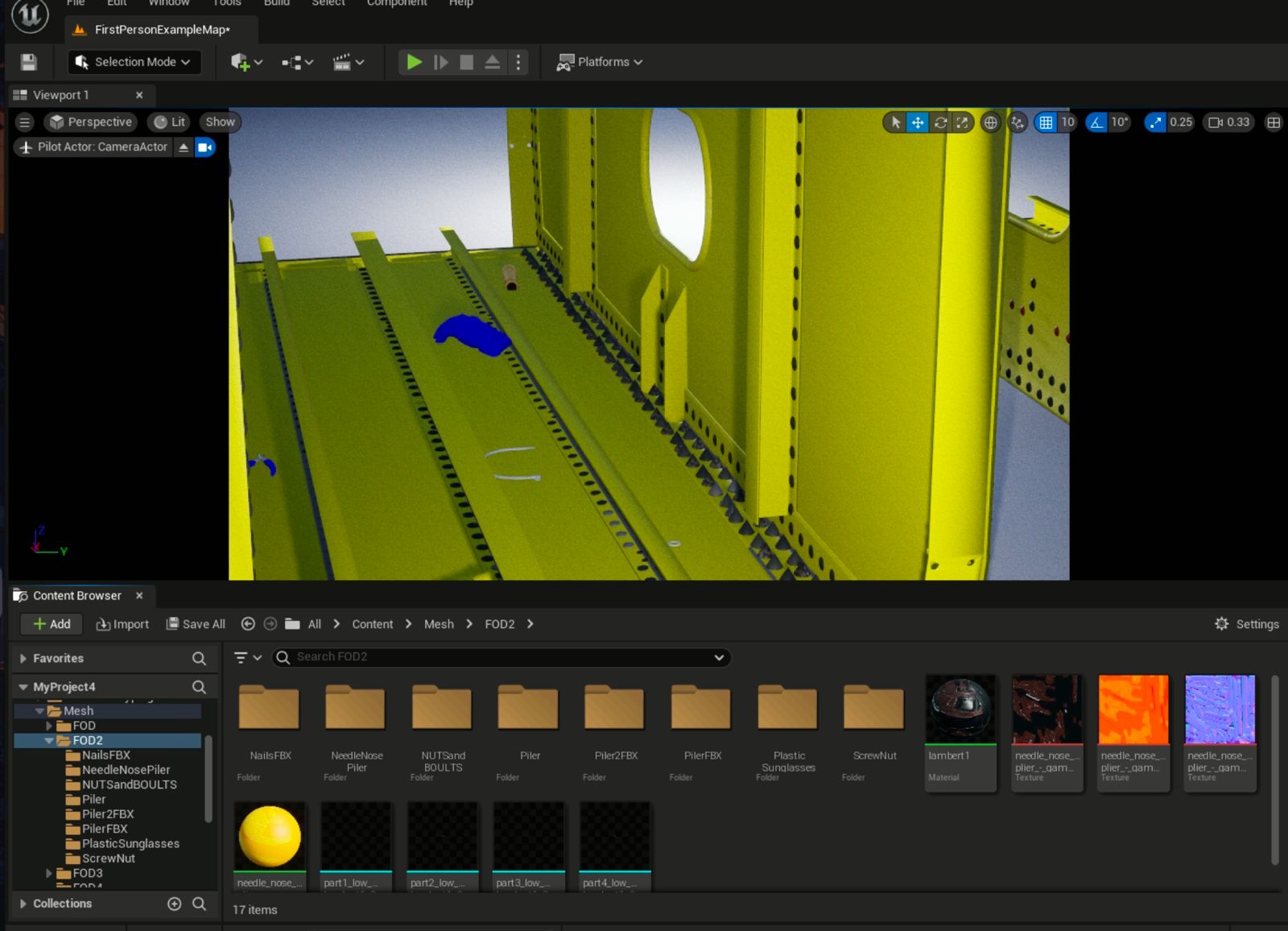}  
    \caption{Unreal Engine editor used to assemble fuel-tank scenes for synthetic FOD data generation, allowing precise control over camera pose, lighting, and object layout.}

    \label{fig:editor}
\end{figure}

\begin{figure}[htbp]
    \centering
    \includegraphics[width=1\linewidth]{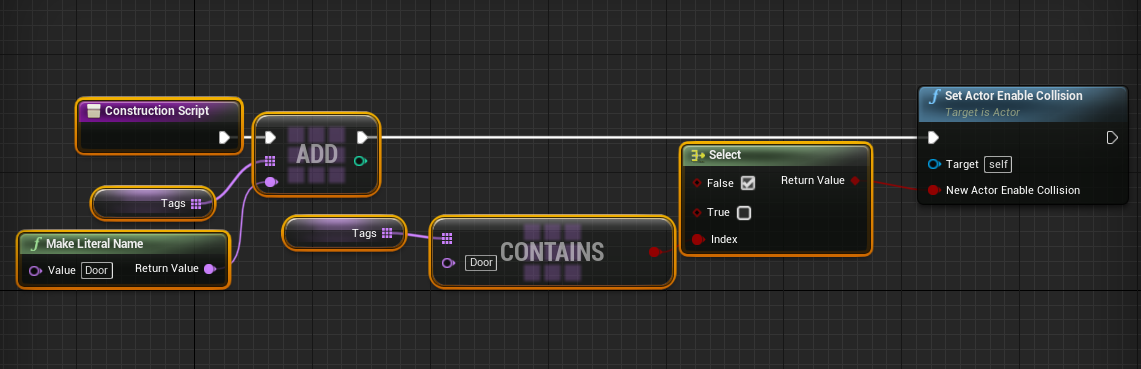}
    \caption{Blueprint scripting for Dataset generation using Unreal Engine.}
    \label{fig:Blueprint}
\end{figure}

Furthermore, the development of robust AI models has been hindered by the lack of comprehensive datasets specifically designed for such complex environments. Most existing FOD datasets focus on outdoor conditions and do not capture the visual and spatial constraints unique to internal aircraft compartments, such as fuel tanks. Collecting and annotating these datasets in aviation settings is labor intensive, expensive, and often restricted by safety regulations. Furthermore, real-world datasets tend to be limited in size and diversity, often under-representing critical FOD classes, which can introduce bias and degrade model performance.
To address these limitations, synthetic data generation through high-fidelity simulations has gained traction as a viable alternative. Using platforms such as Unreal Engine, CARLA, and NVIDIA Isaac Sim\cite{epic_unreal_engine}\cite{Dosovitskiy17}\cite{nvidia_isaac_sim}, researchers can create photorealistic environments \cite{mezghani2020learningvisuallynavigatephotorealistic} and automatically generate richly annotated data, including bounding boxes, semantic masks, and depth information. These simulators provide fine-grained control over variables such as lighting, object placement, and camera perspectives, allowing for the simulation of diverse and rare scenarios.

However, models trained solely on synthetic data often suffer from a domain gap, in which their performance on real-world data is significantly compromised. To bridge this gap, the Sim2Real transfer-learning approach was adopted by researchers \cite{tobin2017domainrandomizationtransferringdeep}. In this method, the model is first trained on a synthetic dataset, and then fine-tuned on a limited real-world dataset to bridge the domain gap and improve its performance in real-world tasks. This approach leverages the efficiency and scalability of synthetic data generation, while ensuring that the model generalizes well to limited real-world scenarios\cite{Horv_th_2023}\cite{prabhu2023bridgingsim2realgapcare}\cite{sadeghi2017sim2realviewinvariantvisual}. This integration leads to enhanced model robustness, reduces reliance on extensive manual annotation, improved generalization, and superior performance when deployed in safety-sensitive applications like FOD detection in aviation.

\begin{figure}
    \centering
    \includegraphics[width=1\linewidth]{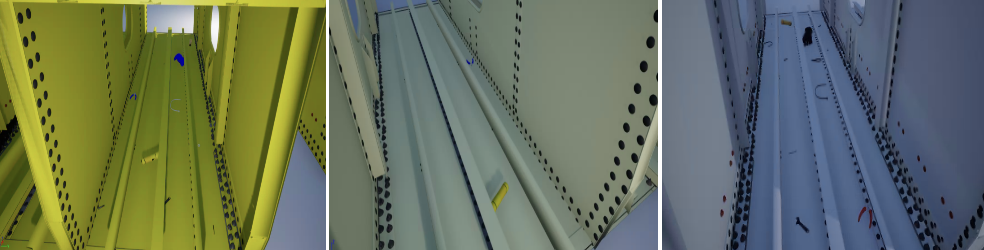}
    \caption{Variation in fuel tank colors within FOD-S2R: the first image depicts a \textbf{yellow} fuel tank, second a \textbf{light yellow} fuel tank, and third a \textbf{gray} fuel tank.}
    \label{fig:color}
\end{figure}

In view of the limitations, FOD-S2R, a novel FOD dataset for Sim2Real based Foreign Object Debris (FOD) is proposed. This dataset includes both synthetic and real-world images. Synthetic data were generated using Unreal Engine to simulate highly realistic fuel tank structures, while the real-world images were acquired from a controlled setup that emulates actual fuel tank environments.
In contrast to most publicly available datasets, which are predominantly captured in open outdoor environments, the FOD-S2R dataset is created within enclosed aviation settings, such as fuel tanks and internal compartments of aircraft. While existing datasets often lack the critical variations necessary for robust FOD detection, including differences in object size, lighting conditions, shadow effects, occlusions, and complex surface textures, the proposed dataset is specifically designed to address these limitations by incorporating such variations through simulation engine. The proposed dataset introduces additional challenges for object detection algorithms owing to the simultaneous presence of multiple objects of varying sizes within visually complex and spatially restricted environments, thereby pushing the limits of current models and promoting the development of more robust and generalizable detection approaches.
The synthetic dataset comprises 3137 images that span 14 classes with approximately 5,509 annotations. The simulated dataset contains various variations such as lighting, distance, field of view, FOD color, fuel tank color (Fig \ref{fig:color}), and object size (Fig \ref{fig:first}), which are very difficult to generate in a real scenario. Similarly, the real-world segment contains 3,114 images with roughly 5,530 annotations across the same classes.

The main contributions of this study are as follows:
\begin{itemize}
    \item  A novel FOD-S2R dataset is introduced for Foreign Object Debris (FOD) detection inside aircraft fuel tanks, comprising real-world images and a  synthetic simulated images generated using Unreal Engine to mimic complex enclosed environments.
    
    \item A benchmark evaluation of state-of-the-art object detection algorithms is conducted on both real and synthetic subsets, highlighting their capabilities and limitations in detecting small, occluded, and spatially scattered FOD under constrained fuel tank conditions.

    \item We study the impact of Sim2Real adaptation techniques on object detection performance, assessing how well synthetic data can improve real-world FOD detection in complex enclosed environments.
     
\end{itemize}

\begin{figure*}[htbp]
\centering
\includegraphics[width=\textwidth, height=0.8\textheight, keepaspectratio]{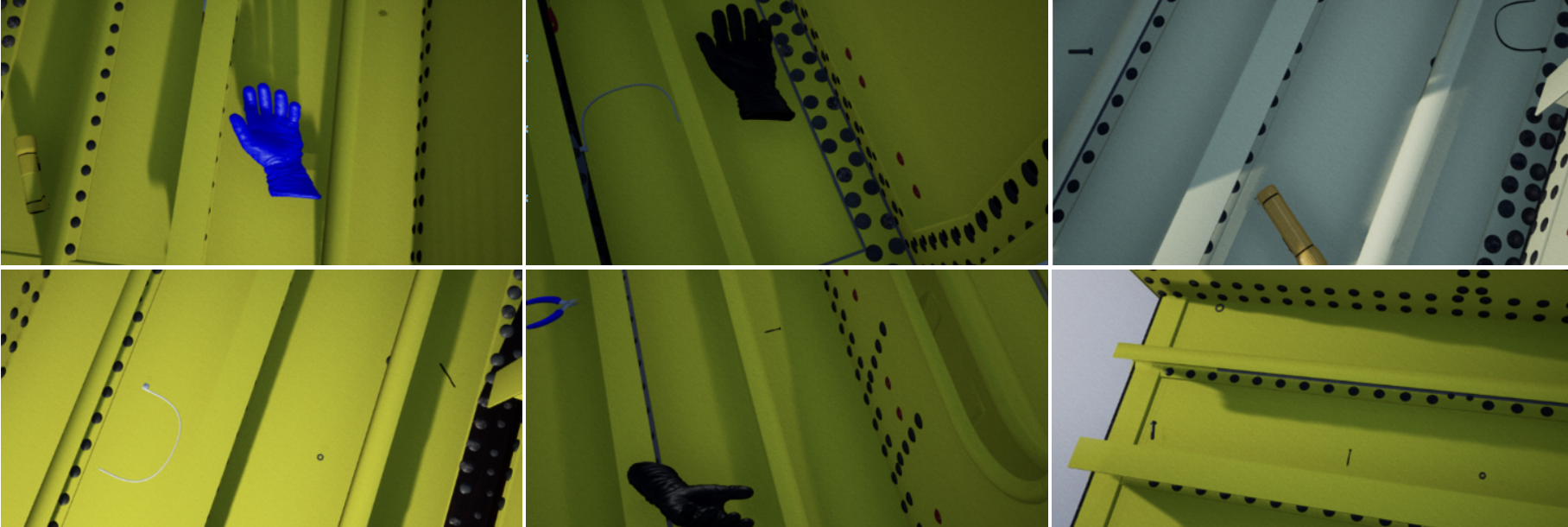}
\caption{The synthetic images in FOD-S2R showcase a range of lighting conditions, diverse glove colors, and variations in the color of the fuel tank, as seen in the  image.}
\label{fig:main image}
\end{figure*}

\begin{figure*}[htbp]
\centering
\includegraphics[width=\textwidth, height=0.8\textheight, keepaspectratio]{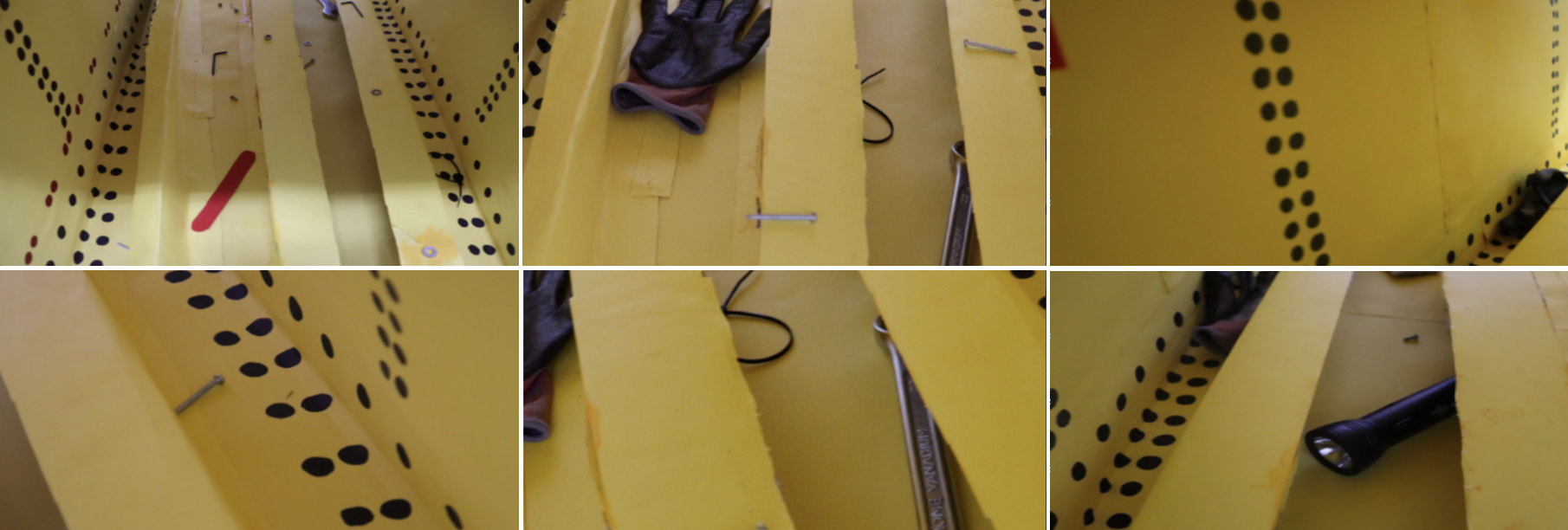}
\caption{The real dataset with varied lightning conditions from different angles. The real dataset includes diverse FOV's, FOD size, lightening conditions etc.}
\label{fig:real}
\end{figure*}

\section{Related Work}

\subsection{Synthetic Data}

Synthetic data have become very famous in modern computer vision domains as they have proven to be a replacement for real datasets or augmented data. Existing simulators\cite{chang2017matterport3dlearningrgbddata,roberts2021hypersimphotorealisticsyntheticdataset} offer controlled, scalable, and cost-effective means to generate large datasets with precise annotations. One of the most influential datasets, ImageNet\cite{5206848}, has significantly influenced synthetic dataset development by serving as a benchmark for object recognition tasks. It contains over 14 million images spanning more than 20,000 categories, and its structure has inspired various synthetic augmentation techniques to improve model generalization. SyntheWorld \cite{song2024syntheworld} is a large-scale synthetic high resolution remote sensing dataset for land cover mapping and building change detection, includes 40,000 images with submeter-level pixels and fine -grained annotations across eight categories, as well as 40,000 pairs of bitemporal images with building change detection. Mechanical Components Benchmark (MCB)\cite{kim2020large} dataset comprises of 3D models of various mechanical components organized into a hierarchical taxonomy based on the International Classification of Standards. Industrial Machine Tool Component Surface Defect Dataset\cite{schlagenhauf2021industrial} - dataset presents images of defects on ball screw drive spindles, capturing the progression of surface defects. SynRS3D \cite{song2025synrs3d}  is another synthetic dataset aimed at global 3D semantic understanding from monocular remote sensing imagery comprising 69,667 high-resolution optical images covering six different city styles worldwide. ShapeNet \cite{chang2015shapenetinformationrich3dmodel} is a large dataset of 3D CAD models over 3000 categories. Multiple datasets offer 3D CAD models precisely aligned with individual objects in real images, facilitating tasks like 3D object recognition and pose estimation \cite{dai2017scannetrichlyannotated3dreconstructions, sun2018pix3ddatasetmethodssingleimage, 6751483, roberts2021hypersimphotorealisticsyntheticdataset, xiang2016objectnet3d , wu2023omniobject3d , zhang2024omni6d}. Other datasets comprise both indoor and outdoor scenes \cite{richter2016playingdatagroundtruth, barua2024gtahdrlargescalesyntheticdataset} GTA  \cite{nikolenko2019syntheticdatadeeplearning, handa2015scenenetunderstandingrealworld, song2016semanticscenecompletionsingle, wu2018buildinggeneralizableagentsrealistic}. 
Most existing synthetic datasets are designed for large-scale land mapping from a bird's-eye view or focus on open environments. While a few efforts have explored synthetic data generation for Foreign Object Debris (FOD) detection, these are often limited in scope, either lacking diversity in object shapes and sizes or failing to account for complex environments such as airplane fuel tanks. In contrast, FOD-S2R includes a wide range of FOD objects with diverse geometries and dimensions, captured under multiple controlled conditions. Notably, the dataset is generated within a confined, enclosed structure that realistically mimics the interior of an aircraft fuel tank, an environment that, to the best of our knowledge, has not been addressed by any existing synthetic dataset.

\subsection{Real Data}
Publicly available detection datasets provide critical benchmarks for FOD-related tasks. However, there are limited datasets that focus on real-world images of small industrial objects. Real-IAD dataset \cite{wang2024realiadrealworldmultiviewdataset}, comprising 150,000 high-resolution images of 30 different industrial objects . For aviation safety, FOD-A\cite{munyer2021fod} introduced the first large-scale real image dataset for FOD detection on airport runways. It contains 31 object classes with over 30000 labeled instances, including metadata for lightning and weather conditions. FOD Material Recognition Dataset\cite{xu2018foreign} an earlier FOD dataset focused on classifying debris by material type containing roughly 3000 instances across 3 categories which does not cover all the types of common FOD's. Although existing real-world FOD datasets are substantial in scope, they are primarily constrained to open environments, such as roads or runways. These settings inherently limit the coverage of rare FOD instances that may occur in confined spaces. Object detection in open environments typically involves minimal background complexity and fewer occlusions, making the task less challenging. In contrast, the FOD-S2R dataset proposed in this work focuses on detecting FOD within the enclosed and visually complex environment of an aircraft fuel tank. This setting introduces significant variability in background textures, colors, and partial occlusions, factors that are critical for developing and evaluating robust object detection models in realistic, safety-critical scenarios.

\subsection{Sim2Real Transfer Learning}
Sim2Real using domain randomaization was proposed by \cite{Horv_th_2023} to bridge the gap between synthetic and real data. ObjectFolder2.0\cite{gao2022objectfolder} comprises 1000 simulated objects with multisensory data, including visual, acoustic, and tactile information. WARM-3D\cite{zhou2024warm} introduces TUMTarf Synthetic dataset and 2D labels from off the shelf detectors for weak supervision in real-world environments. RarePlanes\cite{shermeyer2021rareplanes} proposed  sim2real data of planes from bird eye view contains 253 real scenes of 14,700 aircraft and 50000 synthetic images of 630000 aircrafts. Later \cite{saadiyean2024learning} proposed aerial dataset of roughly 1750 real instances and 2000 simulated instances accross 4 classes: fighter jets, helicopters, carrier planes and passenger planes. Sim2Real-Fire\cite{li2024sim2real} contains 1M virtual and 1 K real wildfire scenarios.
In 2024 \cite{de2024sim} presented real and simulated  multi view images of industrial metal objects. Also \cite{khemmar2022road, bresson2023sim2real} are some other datasets that contain real and simulated images of various environments. 
\subsection{Object Detection}
Object detection is a fundamental task in computer vision that involves identifying and localizing objects in images and videos. It serves as cornerstone for various applications, including autonomous driving, surveillance, and robotics. The advent of deep learning has revolutionized object detection, leading to the development of more robust approaches. RCNN\cite{girshick2014richfeaturehierarchiesaccurate} proposed the first two-stage detector approach by generating region proposals using selective search and CNNs\cite{oshea2015introductionconvolutionalneuralnetworks} to classify these regions. Subsequently, it was improved by \cite{girshick2015fastrcnn,ren2016fasterrcnnrealtimeobject,he2018maskrcnn}. In 2020 YOLO\cite{redmon2016lookonceunifiedrealtime} proposed single regression approach, predicting both class probabilities and bounding boxes simultaneously. These detectors can be classified into two categories: anchor-based\cite{xu2022ppyoloeevolvedversionyolo,li2023yolov6v30fullscalereloading,ge2021yoloxexceedingyoloseries} and anchor-free\cite{bochkovskiy2020yolov4optimalspeedaccuracy, redmon2016yolo9000betterfasterstronger, redmon2018yolov3incrementalimprovement, wang2021scaledyolov4scalingcrossstage, wang2022yolov7trainablebagoffreebiessets }. Further improvements include DDQ\cite{zhang2023densedistinctqueryendtoend} addresses the challenges associated with query design in end-to-end detectors by generating dense queries similar to traditional detectors. Recent advances have demonstrated the incorporation of transformers\cite{vaswani2023attentionneed} into object detection. DETR\cite{carion2020endtoendobjectdetectiontransformers} was the first state-of-the-art model to incorporate transformers to eliminate the need for NMS\cite{hosang2017learningnonmaximumsuppression} and anchor generation. It was improved by \cite{meng2023conditionaldetrfasttraining, zong2023detrscollaborativehybridassignments}and
RT-DETR\cite{zhao2024detrsbeatyolosrealtime}.

\section{Dataset}

\begin{figure}[h]
    \centering
    \includegraphics[width=0.45\textwidth]{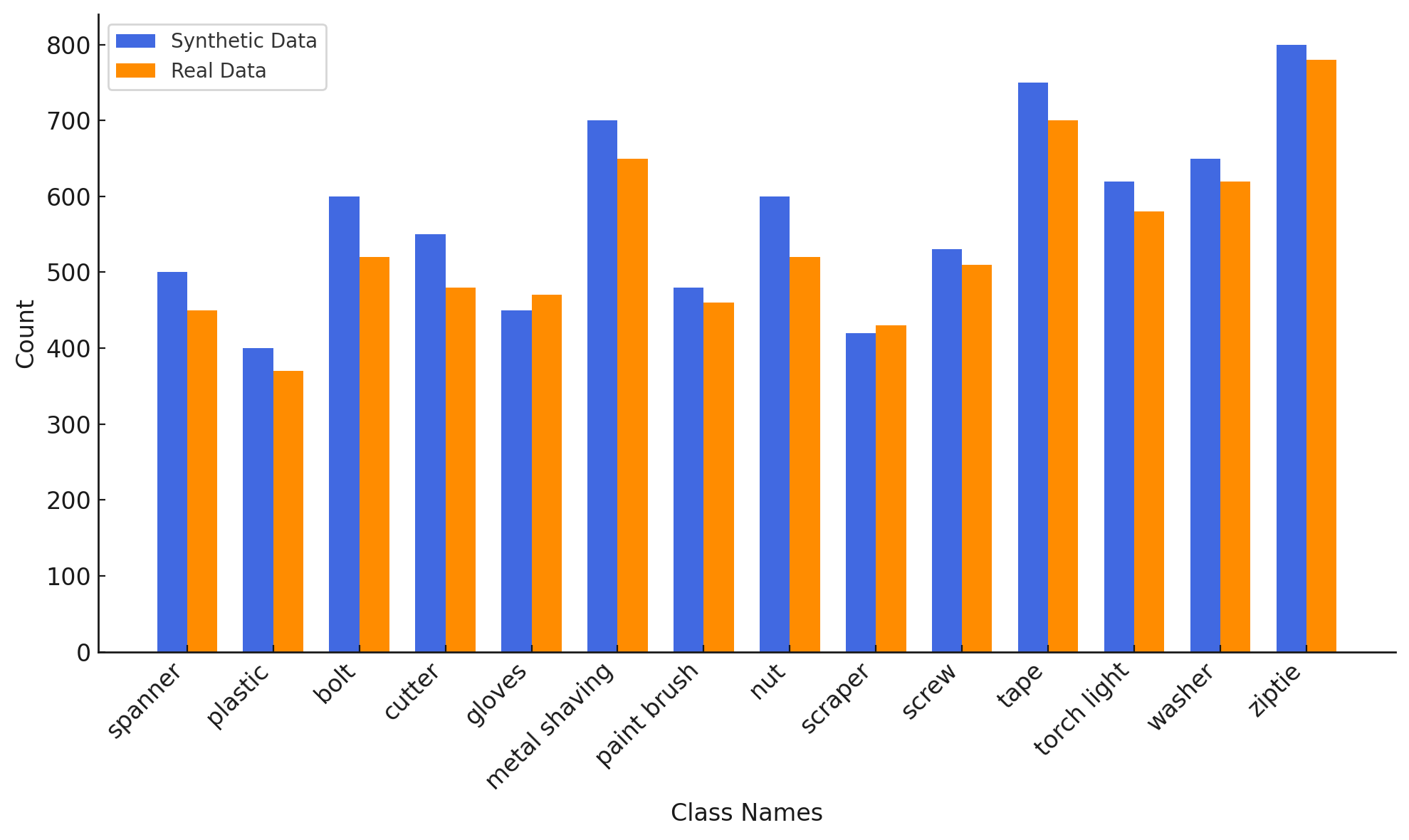}  
    \caption{Instances per category.}
    \label{fig:FOD wise instance}
\end{figure}

In aerospace environments, Foreign Object Debris (FOD) presents a significant challenge, particularly in confined and structurally complex spaces such as aircraft fuel tanks. Unlike open environments such as runways or roads, where the background is relatively plain and uniform, the interior of a fuel tank contains intricate structural components that introduce additional complexity to object detection tasks. These structural elements can create scenarios in which background features resemble FOD, leading to potential misclassification by object detection models. In fuel tanks, various internal features, such as  structural reinforcements, rivets, and surface irregularities, contribute to the challenge of accurately distinguishing the FOD from the internal components of the tank. These structures also create occlusions, where objects may be entirely or partially hidden from view, further complicating detection. To address these challenges, FOD-S2R dataset is proposed.

The FOD-S2R Dataset is a hybrid dataset designed specifically for the detection of Foreign Object Debris (FOD) in aircraft fuel tanks, integrating both real and synthetic image data. The dataset addresses the limitations of existing FOD datasets\cite{munyer2021fod}, which primarily focus on airport runways and lack the environmental complexities of The airplane fuel tank is replicated, including the structural and visual features in both physical and simulated environments, ensuring that the dataset accurately represents the complexities encountered in operational conditions. This dataset enhances generalization and model adaptability for real-world applications in aircraft maintenance.

The FOD-S2R comprises 6,250 high-resolution images, with 3,114 real-world images captured from a custom-built fuel tank replica designed to mimic the structural and lighting conditions of actual aircraft fuel tanks. Real-world images were acquired using a camera under various lighting conditions, ensuring a realistic representation of the operational environment. Additionally, 3,137 synthetic images were generated using the Unreal Engine, incorporating diverse variations in lighting, object placement, occlusion, and reflections to improve the Sim-to-Real  adaptation. Each image is automatically labeled with bounding box annotations, similar to the hierarchical approach used in RarePlanes\cite{shermeyer2021rareplanes}, and further enriched with fine-grained attributes such as object size, texture, occlusion level, and environmental complexity. The distribution of object categories in the dataset is illustrated in Fig \ref{fig:FOD wise instance}.

\subsection{Annotations, Features, and Attributes}
To address these issues, initial manual annotations were used to train a YOLOv11 object detection model, which subsequently generated annotations for the remainder of the dataset. Each model-generated annotation then underwent a rigorous iterative review process. During this review, we carefully verified that each bounding box was properly aligned, non-overlapping, and tightly fitted to the object borders to minimize the inclusion of extraneous background noise or irrelevant features, that could impair detection performance as shown in Fig \ref{fig:first}. This multi-stage hybrid approach-combining automated annotation with comprehensive manual cross-checking-ensures that our annotations are both precise and consistent, providing a robust foundation for subsequent FOD detection research.

\begin{figure}[h]
    \centering
    \includegraphics[width=0.4\textwidth]{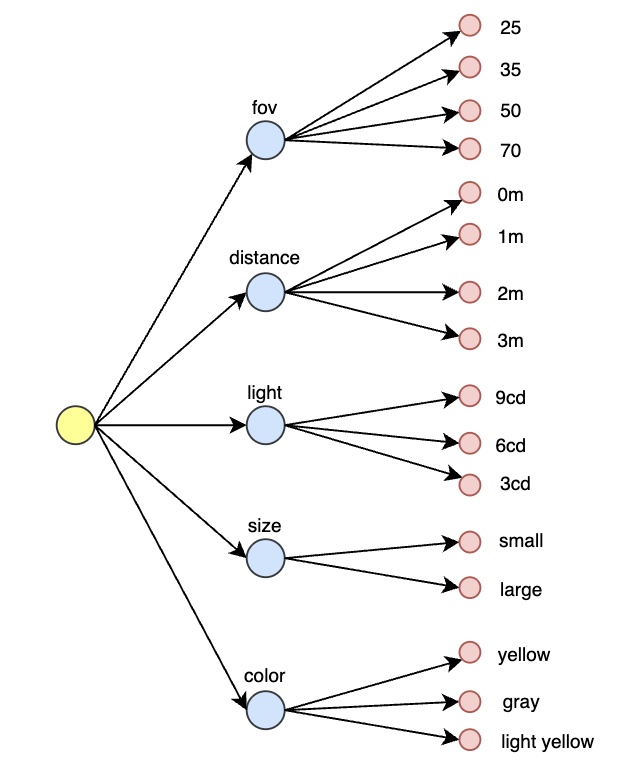}  
    \caption{FOD-S2R dataset showcasing \textbf{5} features and \textbf{16} attributes, allowing exploration of combinations such as field of view, distance, and colour etc.}
    \label{fig:flowchart}
\end{figure}

\subsection{Synthetic Dataset}\label{AA}
\begin{figure}[htbp]
    \centering
    \includegraphics[width=0.5\textwidth]{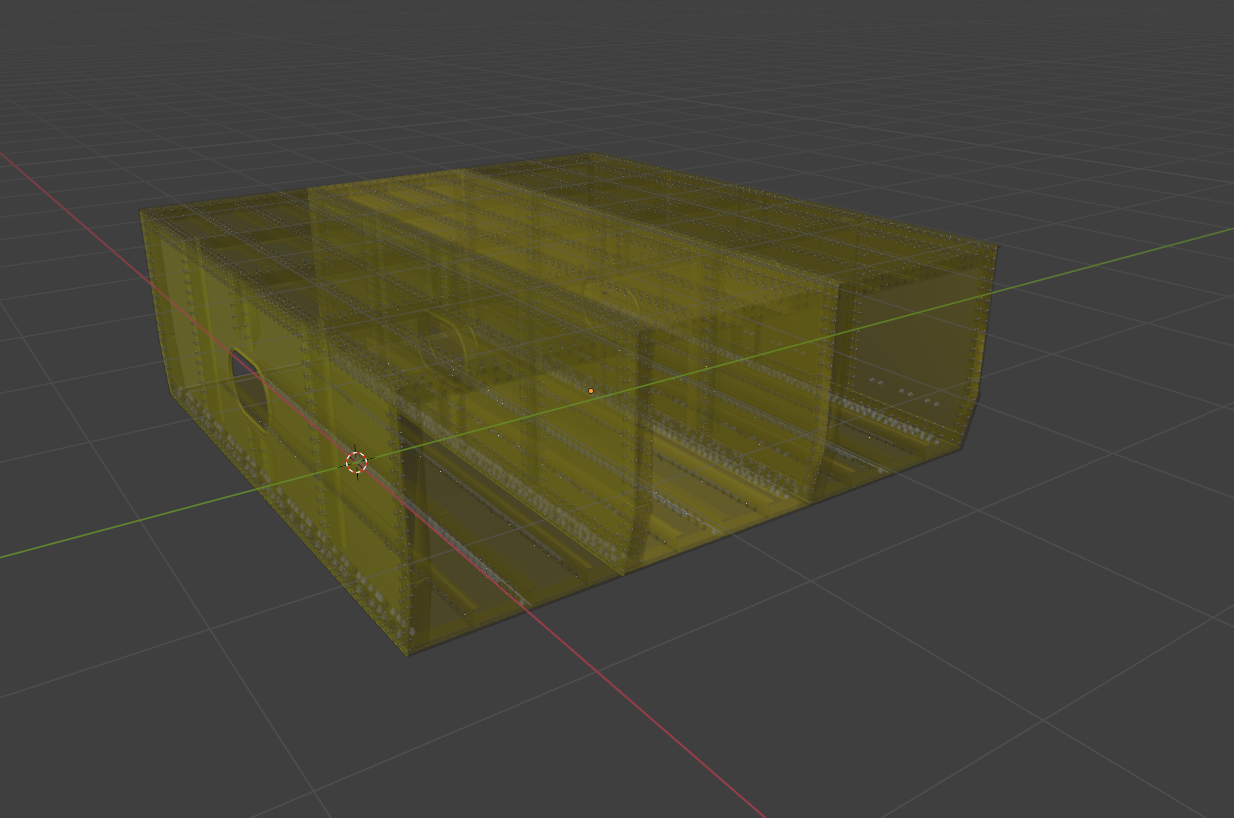}  
    \caption{Aircraft fuel tank inside blender. The fuel tank structure including joints, rivets etc is modeled inside blender and later imported inside Unreal simulation engine.}
    \label{fig:blender}
\end{figure}

The synthetic dataset was generated using a custom simulation environment built on top of Unreal Engine 5\cite{epic_unreal_engine}. The simulated scene was modeled to closely mirror a real aircraft fuel tank structure. The synthetic environment was assembled in Unreal Engine, as shown in \ref{fig:editor}. The annotation and rendering process was automated using Unreal Engine\cite{epic_unreal_engine} Blueprints (Fig. \ref{fig:Blueprint}). The dataset contains 14 classes as mentioned in Fig. \ref{fig:flowchart}.
A detailed CAD model of the actual fuel tank was modeled using Blender, as shown in Fig. \ref{fig:blender}, incorporating real fuel tank dimensions, surface features, and internal complexity. Joints, rivets, and other structural properties were designed to mirror the real fuel tank. Once the model was designed, it was imported into the simulation engine. 

Textures and materials representing real-world conditions were applied within the simulation engine to enhance realism and minimize the domain gap. This includes grease stains, surface irregularities, scratches, and damaged rivets, all of which are commonly observed in the actual fuel tank interior, as shown in Fig. \ref{fig:main image}. Once the fuel tank is completely modeled inside the simulation engine, scene is created.

After scene creation, fuel tank is placed inside the scene. 
The simulation environment, as seen in Fig. \ref{fig:editor}, enables flexible placement of various FOD instances at arbitrary positions within the tank.

\begin{figure}
    \centering
    \includegraphics[width=0.9\linewidth]{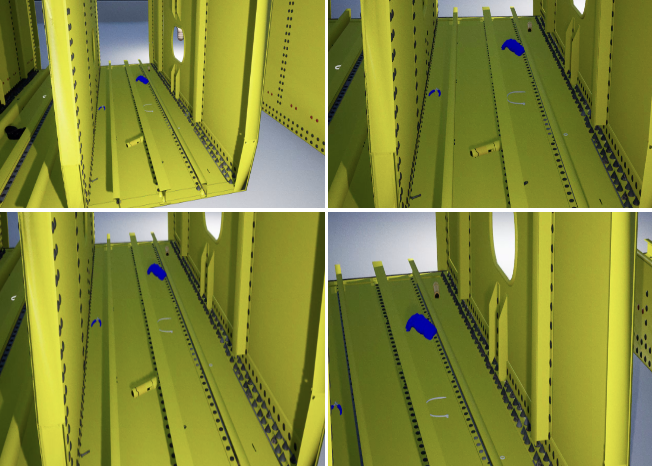}
    \caption{Illustrations showing variations in field of view (FOV) and light intensity. The first image has a \textbf{FOV of 70 and light intensity of 9 cd}, with subsequent images reducing the \textbf{FOV to 25 and light intensity to 3 cd} for comparison.}
    \label{fig:large-small}
\end{figure}

The simulation engine allows user to model the lightening conditions, including light intensity, type etc, as showed in Fig \ref{fig:large-small}
The lighting conditions were carefully adjusted to reflect those observed in actual fuel tank inspections, introducing realistic shadows and illumination gradients. The virtual camera was calibrated to match the physical sensor used in real data collection, replicating parameters such as focal length, resolution, and sensor noise.

To eliminate the need for manual annotation, the Unreal Engine blueprint scripting system was used to automatically generate bounding-box annotations. As seen from Fig. \ref{fig:Blueprint}, the blueprints allow creation of various tags for different object classes. This approach enables the generation of precise annotations tailored to the required format, ensuring compatibility with various object detection frameworks. This automatic data generation approach saves one from time consuming manual annotation.

\subsection{Real Dataset}
Due to the unavailability of a real aircraft fuel tank for image acquisition, a controlled experimental setup was constructed to replicate the structural and environmental conditions typically found within fuel tanks. High-resolution images were captured using a Canon DSLR camera from multiple distances and zoom levels, introducing variation in object scale, field of view, and perspective are showin in Fig \ref{fig:real}. To increase dataset diversity, a video-based image extraction method was also employed, allowing incremental capture of object positions as the camera moved through the tank-like environment.

The setup included structural reinforcements modeled after actual fuel tank features, introducing realistic occlusion challenges. Small FOD items - such as bolts, washers, and nuts-were intentionally placed in partially hidden locations. To capture such occlusions effectively, images were taken from both top-down and horizontal perspectives, adding depth and complexity to the dataset.

To simulate real-world lighting inconsistencies, controlled illumination setups were used, including low-light, direct reflection, and diffuse lighting scenarios. A diverse range of FOD items was incorporated, covering various sizes, shapes, materials, and textures. These included metallic, plastic, rubber, cloth, and electronic components, introducing substantial variation in reflectivity, color, and surface finish, factors that increase the difficulty of accurate object detection under real-world constraints.

\begin{figure}
    \centering
    \includegraphics[width=1\linewidth]{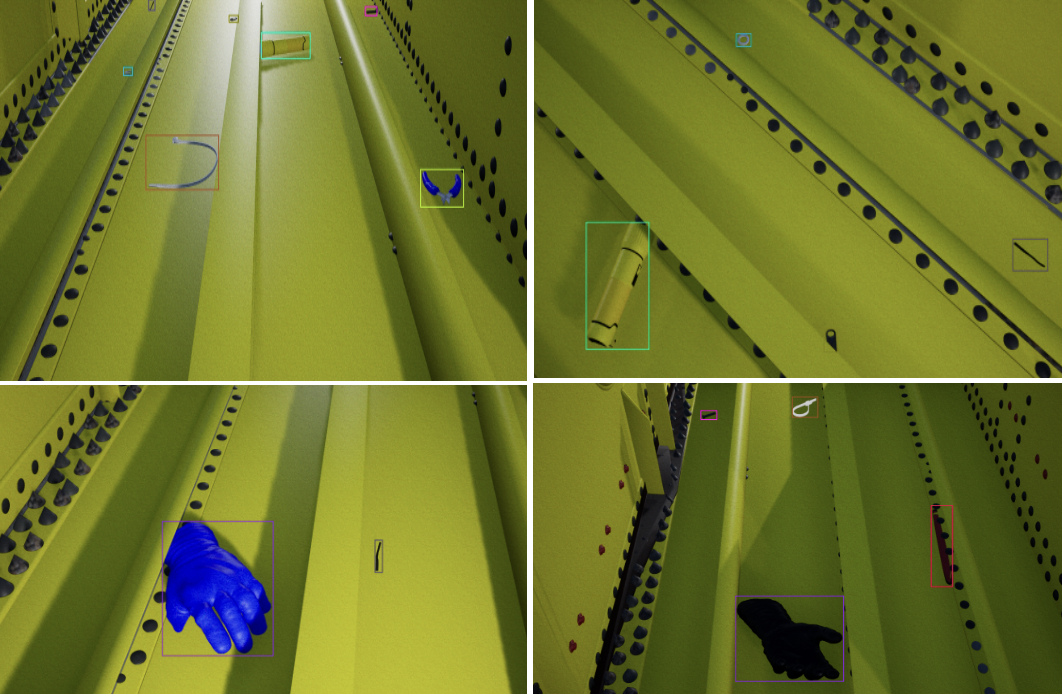}
    \caption{Bounding box annotations focusing on precise localization of small and partially occluded debris, improving model accuracy on confined scenes.}
    \label{fig:first}
\end{figure}

\begin{figure}
    \centering
    \includegraphics[width=1\linewidth]{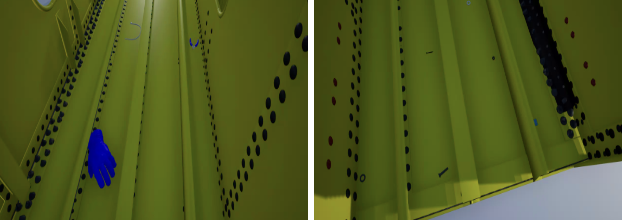}
    \caption{Comparison of larger debris (gloves, cutters, zip ties) and smaller metallic debris (nuts, washers, tape fragments) illustrating object-size imbalance in FOD-S2R.}
    \label{fig:first}
\end{figure}

\section{Experiments}
This section presents a detailed evaluation of FOD-S2R, comprising both real and synthetic images, through a series of object detection experiments. The primary objective of these experiments was to validate the performance of synthetic data in enhancing object detection models and reducing the domain gap between real and simulated images. Specifically, we conducted three experimental configurations to analyze the effectiveness of synthetic data under different training conditions. 

First, we trained the models exclusively on real data to establish a baseline for performance. Second, we evaluated models trained solely on synthetic data to assess their ability to generalize to real-world scenarios. Finally, we explore a fine-tuning approach wherein models are initially pretrained on synthetic data and subsequently fine-tuned with a limited portion of real data. This approach aims to determine whether synthetic data can effectively supplement real-world training and reduce the need for extensive manual annotation.

\subsection{Training and Testing Splits}

To ensure a fair and systematic evaluation, we maintained a consistent 70-10-20 dataset split, where 70\% of the data was used for training, 10\% for validation, and 20\% for testing. This split was applied uniformly across both real and synthetic datasets. Each image, whether real or synthetic, is further divided into 640 × 640 pixel tiles, ensuring that each tile contains at least one instance of Foreign Object Debris (FOD). The dataset was designed to provide a diverse representation of FOD in terms of shape, size, and depth variations, thereby improving the robustness of the model to different object characteristics.

Real-world images in FOD-S2R were acquired using a single camera at a fixed location, ensuring consistency in resolution, lighting conditions, and viewpoint. This controlled acquisition process minimizes variability within the real dataset, making it easier to evaluate the impact of synthetic data. The real images were captured at 25-megapixel resolution, while the synthetic images were generated at 12-megapixel resolution. To maintain uniformity between the two domains, careful preprocessing was applied to ensure consistency in pixel scaling, aspect ratios, and color distributions.

The synthetic dataset was generated with controlled variations in environmental factors, including illumination conditions, object occlusions, and shadowing effects. This variability aims to simulate real-world complexities and enhance the generalization capability of object detection models trained on synthetic data. By incorporating a wide range of environmental conditions, the dataset seeks to mitigate domain adaptation challenges that typically arise when models trained on synthetic data are applied to real-world scenarios.
\\

\begin{table*}[ht]
\centering
\begin{tabular}{lllcccccc}
\toprule
\textbf{MODEL} & \textbf{Category} & \textbf{Dataset} & \textbf{$\text{mAP}_{S}$} & \textbf{$\text{mAP}_{M}$} & \textbf{$\text{mAP}_{L}$} & \textbf{$\text{mAP}_{50}$} & \textbf{$\text{mAP}_{75}$} & \textbf{$\text{mAP}_{50:95}$} \\
\midrule
YOLOv12 & Anchor-Free & syn  &  {\_} &  {\_} &  {\_} & 0.94 & {\_} & 0.718 \\
 &  & real &  {\_} &  {\_} &  &  0.903 & -    & 0.682 \\
\addlinespace
YOLOv11 & Anchor-Free & syn  & 0.670 & 0.786 & 0.450 & 0.845 & {\_} & 0.568 \\
 &  & real & 0.743 & 0.517 & 0.553 & 0.929 & -    & 0.715 \\
\addlinespace
YOLOv8  & Anchor-Free & syn  & 0.645 & 0.787 & 0.443 & 0.853 & {\_} & 0.572 \\
  &  & real & 0.713 & 0.510 & 0.492 & 0.913 & -    & 0.707 \\
\addlinespace
RT-DETR & Anchor-Free & syn  & 0.640 & 0.570 & 0.610 & 0.815 & - & 0.546 \\
 &  & real & 0.601 & 0.460 & 0.570 & 0.917 & - & 0.685 \\
\addlinespace
DDQ & Anchor-Free & syn  & 0.285 & 0.314 & 0.340 & 0.584 & 0.359 & 0.475 \\
 &  & real &  0.185 & 0.227 & 0.250 & 0.471 & 0.341 & 0.360 \\
\addlinespace

\addlinespace

YOLOv5  & Anchor-Based & syn  & 0.670 & 0.788 & 0.465 & 0.845 & - & 0.566 \\
   & & real & 0.696 & 0.518 & 0.515 & 0.923 & --    & 0.708 \\
\addlinespace
Faster R-CNN & Anchor-Based & syn  & 0.552 & 0.280 & 0.261 & 0.831 & 0.420 & 0.440 \\
 &  & real & 0.226 & 0.583 & 0.818 & 0.848 & 0.659 & 0.594 \\
\addlinespace
RetinaNet    & Anchor-Based & syn  & 0.516 & 0.735 & 0.852 & 0.894 & 0.772 & 0.650 \\
    &  & real & 0.235 & 0.651 & 0.781 & 0.879 & 0.705 & 0.637 \\
\addlinespace
RF-DETR    & Anchor-Based & syn  & 0.441 & 0.702 & 0.883 & 0.843 & 0.597 & 0.550 \\
    &  & real & 0.383 & 0.700 & 0.887 & 0.930 & 0.782 & 0.702 \\
\bottomrule
\end{tabular}
\caption{Model performance comparison on synthetic and real datasets.}
\label{tab:baseline_results}
\end{table*}
% Please add the following required packages to your document preamble:
% \usepackage{booktabs}

\section{Implementation}

To evaluate the performance of the proposed dataset across diverse object detection paradigms, experiments were carried out using both anchor-based and anchor-free models. The anchor-based category included YOLOv3\cite{redmon2018yolov3incrementalimprovement}, Faster R-CNN\cite{ren2016fasterrcnnrealtimeobject}, and Mask R-CNN\cite{he2018maskrcnn}, which rely on predefined anchor templates for bounding box regression. In contrast, anchor-free architectures such as YOLOv12, YOLOv11, DDQ\cite{zhang2023densedistinctqueryendtoend}, and transformer-based detectors including RT-DETR\cite{zhao2024detrsbeatyolosrealtime} eliminate anchor mechanisms and instead formulate object localization either through point-based regression or attention-guided query decoding. This model diversity enables a controlled comparison between conventional anchor-driven detection pipelines and emerging transformer and query-based approaches.

A ResNet-50 backbone\cite{he2015deepresiduallearningimage} pretrained on ImageNet\cite{5206848} was used uniformly across all detectors to maintain consistent representation learning capability and minimize initialization variance. All models were implemented using the Detectron2 framework\cite{wu2019detectron2} and executed on two NVIDIA RTX 2080 GPUs equipped with 12\,GB memory per card. Image augmentation was applied uniformly to synthetic and real subsets, including saturation perturbation between $-20\%$ and $+20\%$, ensuring robustness to lighting variation without introducing domain-specific artifacts. 

All experiments were trained for 100 epochs with a fixed learning rate of 0.001, batch size of 1, and default optimization and scheduling parameters, ensuring consistent computational exposure across architectures and enabling fair comparative evaluation.

\begin{table*}[ht]
\centering

\begin{tabular}{llcccccccc}
\toprule
\textbf{Model} & \textbf{Train} & \textbf{Fine-Tune (FT)} & \textbf{Test} & 
\textbf{$\text{mAP}_{S}$} & \textbf{$\text{mAP}_{M}$} & \textbf{$\text{mAP}_{L}$} & \textbf{$\text{mAP}_{50}$} & \textbf{$\text{mAP}_{75}$} & \textbf{$\text{mAP}_{50:95}$} \\
\midrule
RF-DETR & real & syn & real & 0.676 & 0.760 & 0.872 & 0.931 & 0.782 & 0.740 \\
        & syn & real & real & 0.560 & 0.723 & 0.845 & 0.913 & 0.750 & 0.712 \\
\addlinespace
Yolov12 & Real & Synthetic & Real & {\_}  &  {\_} & {\_}  & 0.898 &{\_}  & 0.679 \\
      & Synthetic & Real & Real & {\_} & {\_}  & {\_}  & 0.902 &  {\_} & 0.69 \\
\bottomrule
\end{tabular}
\caption{Effect of training and fine-tuning order on model performance.}
\label{tab:sim2real}
\end{table*}

\subsection{Results and Discussion}

The performance of all evaluated models on both synthetic and real subsets of the FOD-S2R dataset is summarized in Table~\ref{tab:baseline_results}. Evaluation was conducted using COCO and YOLO metrics, with emphasis on $\text{mAP}_{50}$, $\text{mAP}_{75}$, and $\text{mAP}_{50:95}$, along with scale-aware metrics $\text{mAP}_{S}$, $\text{mAP}_{M}$, and $\text{mAP}_{L}$. The results reveal a clear domain disparity across models. In most cases, detectors trained and evaluated on real imagery outperform those tested on synthetic samples, particularly under stricter localization thresholds. For instance, YOLOv11 reports an $\text{mAP}_{50}$ of 0.845 on synthetic data and improves to 0.929 on real images, while its $\text{mAP}_{50:95}$ increases from 0.568 to 0.715. YOLOv5 and YOLOv8 exhibit similar behavior, indicating that real samples contain more detectable structural cues, especially for coarse bounding box alignment.

Small object detection presents a recurrent challenge. Although some architectures show better real-domain localization at coarse IoU thresholds, scale analysis exposes reduced robustness. RF-DETR demonstrates the highest real-domain performance in overall detection, reaching $\text{mAP}_{50}=0.930$ and $\text{mAP}_{50:95}=0.702$, yet its $\text{mAP}_{S}=0.383$ indicates degraded sensitivity to small debris, such as washers or cable ties. A contrasting pattern appears in transformer-based RT-DETR. While it performs strongly on synthetic data with $\text{mAP}_{S}=0.640$, its small-scale performance declines when evaluated on real imagery ($\text{mAP}_{S}=0.601$), suggesting increased susceptibility to real texture variance and illumination noise despite strong global feature modeling.

These observations support the existence of a measurable domain gap between synthetic and real conditions. Synthetic data, despite being geometrically accurate, remains more uniform in lighting, surface reflectance, and occlusion patterns. Real-world data introduce significantly higher variation due to shadow artifacts, paint degradation, and unpredictable object pose, contributing to reduced localization precision at higher IoU thresholds. This discrepancy is most evident in $\text{mAP}_{50:95}$ metrics, where several models display drops of 10--20\% when transitioning between domains, confirming that generalization without adaptation remains nontrivial.

\subsection{Sim2Real Experiments}

To investigate whether this performance gap can be mitigated, domain adaptation experiments were conducted using RF-DETR and YOLOv12 as the baseline model, owing to its strong real-domain results in table \ref{tab:sim2real}. Two adaptation strategies were evaluated: synthetic pretraining followed by fine-tuning on real data, and the reverse configuration. When pretrained on synthetic samples and fine-tuned on real images, the model achieved its highest performance on the real test set, reaching $\text{mAP}_{50}=0.931$ and $\text{mAP}_{50:95}=0.740$. Notably, small-object performance also improved significantly from $\text{mAP}_{S}=0.383$ in the baseline real-only condition to $\text{mAP}_{S}=0.676$, indicating that synthetic variability provides a useful prior for learning scale-invariant localization cues.

In contrast, reversing the adaptation sequence (real-first, synthetic-second) resulted in weaker performance, with $\text{mAP}_{50:95}$ dropping to 0.712 despite a comparable $\text{mAP}_{50}$ of 0.913. This behavior suggests that once the model converges on features present in real imagery, exposure to synthetic textures may degrade previously learned domain-specific representations. The synthetic domain, while controlled and photorealistic, lacks the high-frequency surface irregularities found in real fuel tank environments, preventing the model from fully retaining fine-grained recognition ability during the second stage of training.

The results indicate that synthetic data are most effective when used as a pretraining resource rather than as a replacement for or refinement stage after real data. Fine-tuning on real samples remains essential for final convergence, while synthetic-first training provides wider distribution coverage and strengthens generalization capacity under previously unseen viewpoints and object configurations. These findings reinforce the viability of synthetic data as a scalable initialization strategy for enclosed industrial inspection tasks, particularly when real annotated data are limited.

\section{Conclusion}
This work addresses a critical gap in aviation safety by introducing FOD-S2R, a novel hybrid dataset for the detection of foreign objects (FOD) inside aircraft fuel tanks. By combining limited real-world images with diverse synthetic samples generated using Unreal Engine, the dataset enables robust evaluation of object detection models under challenging fuel tank conditions. 

A comprehensive benchmarking study was conducted across a range of anchor-based and anchor-free object detection models, incorporating both fine-tuning and retraining using real and synthetic datasets. This benchmarking highlights architectural limitations in handling small, occluded, and visually diverse FOD items in cluttered environments.
Further Sim2Real transfer learning experiments demonstrate that integrating synthetic data improves detection performance by 15\%, reducing the dependency on costly and limited real-world data.
Thus, this work supports the potential to reduce the dependence on labor-intensive real-world data collection by leveraging synthetic data, thus contributing to the development of more efficient and reliable automated FOD detection systems for aviation maintenance.

%\cite{b6}.
\bibliography{fod}

\begin{thebibliography}{10}

\bibitem{barua2024gtahdrlargescalesyntheticdataset}
Hrishav~Bakul Barua, Kalin Stefanov, KokSheik Wong, Abhinav Dhall, and Ganesh Krishnasamy.
\newblock Gta-hdr: A large-scale synthetic dataset for hdr image reconstruction, 2024.

\bibitem{bochkovskiy2020yolov4optimalspeedaccuracy}
Alexey Bochkovskiy, Chien-Yao Wang, and Hong-Yuan~Mark Liao.
\newblock Yolov4: Optimal speed and accuracy of object detection, 2020.

\bibitem{bresson2023sim2real}
Marc Bresson, Yang Xing, and Weisi Guo.
\newblock Sim2real: Generative ai to enhance photorealism through domain transfer with gan and seven-chanel-360°-paired-images dataset.
\newblock {\em Sensors}, 24(1):94, 2023.

\bibitem{carion2020endtoendobjectdetectiontransformers}
Nicolas Carion, Francisco Massa, Gabriel Synnaeve, Nicolas Usunier, Alexander Kirillov, and Sergey Zagoruyko.
\newblock End-to-end object detection with transformers, 2020.

\bibitem{chang2017matterport3dlearningrgbddata}
Angel Chang, Angela Dai, Thomas Funkhouser, Maciej Halber, Matthias Nießner, Manolis Savva, Shuran Song, Andy Zeng, and Yinda Zhang.
\newblock Matterport3d: Learning from rgb-d data in indoor environments, 2017.

\bibitem{chang2015shapenetinformationrich3dmodel}
Angel~X. Chang, Thomas Funkhouser, Leonidas Guibas, Pat Hanrahan, Qixing Huang, Zimo Li, Silvio Savarese, Manolis Savva, Shuran Song, Hao Su, Jianxiong Xiao, Li~Yi, and Fisher Yu.
\newblock Shapenet: An information-rich 3d model repository, 2015.

\bibitem{dai2017scannetrichlyannotated3dreconstructions}
Angela Dai, Angel~X. Chang, Manolis Savva, Maciej Halber, Thomas Funkhouser, and Matthias Nießner.
\newblock Scannet: Richly-annotated 3d reconstructions of indoor scenes, 2017.

\bibitem{de2024sim}
Peter De~Roovere, Steven Moonen, Nick Michiels, and Francis Wyffels.
\newblock Sim-to-real dataset of industrial metal objects.
\newblock {\em Machines}, 12(2):99, 2024.

\bibitem{5206848}
Jia Deng, Wei Dong, Richard Socher, Li-Jia Li, Kai Li, and Li~Fei-Fei.
\newblock Imagenet: A large-scale hierarchical image database.
\newblock In {\em 2009 IEEE Conference on Computer Vision and Pattern Recognition}, pages 248--255, 2009.

\bibitem{Dosovitskiy17}
Alexey Dosovitskiy, German Ros, Felipe Codevilla, Antonio Lopez, and Vladlen Koltun.
\newblock Carla: An open urban driving simulator.
\newblock In {\em Proceedings of the 1st Annual Conference on Robot Learning (CoRL)}, pages 1--16, 2017.

\bibitem{farooq2024improved}
Javaria Farooq, Muhammad Muaz, Khurram Khan~Jadoon, Nayyer Aafaq, and Muhammad Khizer~Ali Khan.
\newblock An improved yolov8 for foreign object debris detection with optimized architecture for small objects.
\newblock {\em Multimedia Tools and Applications}, 83(21):60921--60947, 2024.

\bibitem{epic_unreal_engine}
Epic Games.
\newblock Unreal engine, n.d.
\newblock Accessed: Apr. 15, 2025.

\bibitem{gao2022objectfolder}
Ruohan Gao, Zilin Si, Yen-Yu Chang, Samuel Clarke, Jeannette Bohg, Li~Fei-Fei, Wenzhen Yuan, and Jiajun Wu.
\newblock Objectfolder 2.0: A multisensory object dataset for sim2real transfer.
\newblock In {\em Proceedings of the IEEE/CVF conference on computer vision and pattern recognition}, pages 10598--10608, 2022.

\bibitem{ge2021yoloxexceedingyoloseries}
Zheng Ge, Songtao Liu, Feng Wang, Zeming Li, and Jian Sun.
\newblock Yolox: Exceeding yolo series in 2021, 2021.

\bibitem{girshick2015fastrcnn}
Ross Girshick.
\newblock Fast r-cnn, 2015.

\bibitem{girshick2014richfeaturehierarchiesaccurate}
Ross Girshick, Jeff Donahue, Trevor Darrell, and Jitendra Malik.
\newblock Rich feature hierarchies for accurate object detection and semantic segmentation, 2014.

\bibitem{handa2015scenenetunderstandingrealworld}
Ankur Handa, Viorica Patraucean, Vijay Badrinarayanan, Simon Stent, and Roberto Cipolla.
\newblock Scenenet: Understanding real world indoor scenes with synthetic data, 2015.

\bibitem{he2018maskrcnn}
Kaiming He, Georgia Gkioxari, Piotr Dollár, and Ross Girshick.
\newblock Mask r-cnn, 2018.

\bibitem{he2015deepresiduallearningimage}
Kaiming He, Xiangyu Zhang, Shaoqing Ren, and Jian Sun.
\newblock Deep residual learning for image recognition, 2015.

\bibitem{Horv_th_2023}
Dániel Horváth, Gábor Erdős, Zoltán Istenes, Tomáš Horváth, and Sándor Földi.
\newblock Object detection using sim2real domain randomization for robotic applications.
\newblock {\em IEEE Transactions on Robotics}, 39(2):1225–1243, April 2023.

\bibitem{hosang2017learningnonmaximumsuppression}
Jan Hosang, Rodrigo Benenson, and Bernt Schiele.
\newblock Learning non-maximum suppression, 2017.

\bibitem{khemmar2022road}
Redouane Khemmar, Antoine Mauri, Camille Dulompont, Jayadeep Gajula, Vincent Vauchey, Madjid Haddad, and R{\'e}mi Boutteau.
\newblock Road and railway smart mobility: A high-definition ground truth hybrid dataset.
\newblock {\em Sensors}, 22(10):3922, 2022.

\bibitem{kim2020large}
Sangpil Kim, Hyung-gun Chi, Xiao Hu, Qixing Huang, and Karthik Ramani.
\newblock A large-scale annotated mechanical components benchmark for classification and retrieval tasks with deep neural networks.
\newblock In {\em Computer Vision--ECCV 2020: 16th European Conference, Glasgow, UK, August 23--28, 2020, Proceedings, Part XVIII 16}, pages 175--191. Springer, 2020.

\bibitem{li2023yolov6v30fullscalereloading}
Chuyi Li, Lulu Li, Yifei Geng, Hongliang Jiang, Meng Cheng, Bo~Zhang, Zaidan Ke, Xiaoming Xu, and Xiangxiang Chu.
\newblock Yolov6 v3.0: A full-scale reloading, 2023.

\bibitem{li2024sim2real}
Yanzhi Li, Keqiu Li, LI~GUOHUI, Chanqing Ji, Lubo Wang, Die Zuo, Qing Guo, Feng Zhang, Manyu Wang, Di~Lin, et~al.
\newblock Sim2real-fire: A multi-modal simulation dataset for forecast and backtracking of real-world forest fire.
\newblock {\em Advances in Neural Information Processing Systems}, 37:1428--1442, 2024.

\bibitem{6751483}
Joseph~J. Lim, Hamed Pirsiavash, and Antonio Torralba.
\newblock Parsing ikea objects: Fine pose estimation.
\newblock In {\em 2013 IEEE International Conference on Computer Vision}, pages 2992--2999, 2013.

\bibitem{meng2023conditionaldetrfasttraining}
Depu Meng, Xiaokang Chen, Zejia Fan, Gang Zeng, Houqiang Li, Yuhui Yuan, Lei Sun, and Jingdong Wang.
\newblock Conditional detr for fast training convergence, 2023.

\bibitem{mezghani2020learningvisuallynavigatephotorealistic}
Lina Mezghani, Sainbayar Sukhbaatar, Arthur Szlam, Armand Joulin, and Piotr Bojanowski.
\newblock Learning to visually navigate in photorealistic environments without any supervision, 2020.

\bibitem{munyer2021fod}
Travis Munyer, Pei-Chi Huang, Chenyu Huang, and Xin Zhong.
\newblock Fod-a: A dataset for foreign object debris in airports.
\newblock {\em arXiv preprint arXiv:2110.03072}, 2021.

\bibitem{nikolenko2019syntheticdatadeeplearning}
Sergey~I. Nikolenko.
\newblock Synthetic data for deep learning, 2019.

\bibitem{nvidia_isaac_sim}
NVIDIA.
\newblock Isaac sim, n.d.
\newblock Accessed: Apr. 15, 2025.

\bibitem{oshea2015introductionconvolutionalneuralnetworks}
Keiron O'Shea and Ryan Nash.
\newblock An introduction to convolutional neural networks, 2015.

\bibitem{prabhu2023bridgingsim2realgapcare}
Viraj Prabhu, David Acuna, Andrew Liao, Rafid Mahmood, Marc~T. Law, Judy Hoffman, Sanja Fidler, and James Lucas.
\newblock Bridging the sim2real gap with care: Supervised detection adaptation with conditional alignment and reweighting, 2023.

\bibitem{redmon2016lookonceunifiedrealtime}
Joseph Redmon, Santosh Divvala, Ross Girshick, and Ali Farhadi.
\newblock You only look once: Unified, real-time object detection, 2016.

\bibitem{redmon2016yolo9000betterfasterstronger}
Joseph Redmon and Ali Farhadi.
\newblock Yolo9000: Better, faster, stronger, 2016.

\bibitem{redmon2018yolov3incrementalimprovement}
Joseph Redmon and Ali Farhadi.
\newblock Yolov3: An incremental improvement, 2018.

\bibitem{ren2016fasterrcnnrealtimeobject}
Shaoqing Ren, Kaiming He, Ross Girshick, and Jian Sun.
\newblock Faster r-cnn: Towards real-time object detection with region proposal networks, 2016.

\bibitem{richter2016playingdatagroundtruth}
Stephan~R. Richter, Vibhav Vineet, Stefan Roth, and Vladlen Koltun.
\newblock Playing for data: Ground truth from computer games, 2016.

\bibitem{roberts2021hypersimphotorealisticsyntheticdataset}
Mike Roberts, Jason Ramapuram, Anurag Ranjan, Atulit Kumar, Miguel~Angel Bautista, Nathan Paczan, Russ Webb, and Joshua~M. Susskind.
\newblock Hypersim: A photorealistic synthetic dataset for holistic indoor scene understanding, 2021.

\bibitem{saadiyean2024learning}
Qiranul Saadiyean, SP~Samprithi, and Suresh Sundaram.
\newblock Learning multi-scale context mask-rcnn network for slant angled aerial imagery in instance segmentation in a sim2real setup.
\newblock In {\em 2024 IEEE International Conference on Robotics and Automation (ICRA)}, pages 13573--13580. IEEE, 2024.

\bibitem{sadeghi2017sim2realviewinvariantvisual}
Fereshteh Sadeghi, Alexander Toshev, Eric Jang, and Sergey Levine.
\newblock Sim2real view invariant visual servoing by recurrent control, 2017.

\bibitem{schlagenhauf2021industrial}
Tobias Schlagenhauf and Magnus Landwehr.
\newblock Industrial machine tool component surface defect dataset.
\newblock {\em Data in Brief}, 39:107643, 2021.

\bibitem{shermeyer2021rareplanes}
Jacob Shermeyer, Thomas Hossler, Adam Van~Etten, Daniel Hogan, Ryan Lewis, and Daeil Kim.
\newblock Rareplanes: Synthetic data takes flight.
\newblock In {\em Proceedings of the IEEE/CVF Winter Conference on Applications of Computer Vision}, pages 207--217, 2021.

\bibitem{song2025synrs3d}
Jian Song, Hongruixuan Chen, Weihao Xuan, Junshi Xia, and Naoto Yokoya.
\newblock Synrs3d: A synthetic dataset for global 3d semantic understanding from monocular remote sensing imagery.
\newblock {\em Advances in Neural Information Processing Systems}, 37:117388--117425, 2025.

\bibitem{song2024syntheworld}
Jian Song, Hongruixuan Chen, and Naoto Yokoya.
\newblock Syntheworld: A large-scale synthetic dataset for land cover mapping and building change detection.
\newblock In {\em Proceedings of the IEEE/CVF Winter Conference on Applications of Computer Vision}, pages 8287--8296, 2024.

\bibitem{song2016semanticscenecompletionsingle}
Shuran Song, Fisher Yu, Andy Zeng, Angel~X. Chang, Manolis Savva, and Thomas Funkhouser.
\newblock Semantic scene completion from a single depth image, 2016.

\bibitem{sun2018pix3ddatasetmethodssingleimage}
Xingyuan Sun, Jiajun Wu, Xiuming Zhang, Zhoutong Zhang, Chengkai Zhang, Tianfan Xue, Joshua~B. Tenenbaum, and William~T. Freeman.
\newblock Pix3d: Dataset and methods for single-image 3d shape modeling, 2018.

\bibitem{tobin2017domainrandomizationtransferringdeep}
Josh Tobin, Rachel Fong, Alex Ray, Jonas Schneider, Wojciech Zaremba, and Pieter Abbeel.
\newblock Domain randomization for transferring deep neural networks from simulation to the real world, 2017.

\bibitem{vaswani2023attentionneed}
Ashish Vaswani, Noam Shazeer, Niki Parmar, Jakob Uszkoreit, Llion Jones, Aidan~N. Gomez, Lukasz Kaiser, and Illia Polosukhin.
\newblock Attention is all you need, 2023.

\bibitem{wang2024realiadrealworldmultiviewdataset}
Chengjie Wang, Wenbing Zhu, Bin-Bin Gao, Zhenye Gan, Jianning Zhang, Zhihao Gu, Shuguang Qian, Mingang Chen, and Lizhuang Ma.
\newblock Real-iad: A real-world multi-view dataset for benchmarking versatile industrial anomaly detection, 2024.

\bibitem{wang2021scaledyolov4scalingcrossstage}
Chien-Yao Wang, Alexey Bochkovskiy, and Hong-Yuan~Mark Liao.
\newblock Scaled-yolov4: Scaling cross stage partial network, 2021.

\bibitem{wang2022yolov7trainablebagoffreebiessets}
Chien-Yao Wang, Alexey Bochkovskiy, and Hong-Yuan~Mark Liao.
\newblock Yolov7: Trainable bag-of-freebies sets new state-of-the-art for real-time object detectors, 2022.

\bibitem{wu2023omniobject3d}
Tong Wu, Jiarui Zhang, Xiao Fu, Yuxin Wang, Jiawei Ren, Liang Pan, Wayne Wu, Lei Yang, Jiaqi Wang, Chen Qian, et~al.
\newblock Omniobject3d: Large-vocabulary 3d object dataset for realistic perception, reconstruction and generation.
\newblock In {\em Proceedings of the IEEE/CVF Conference on Computer Vision and Pattern Recognition}, pages 803--814, 2023.

\bibitem{wu2018buildinggeneralizableagentsrealistic}
Yi~Wu, Yuxin Wu, Georgia Gkioxari, and Yuandong Tian.
\newblock Building generalizable agents with a realistic and rich 3d environment, 2018.

\bibitem{wu2019detectron2}
Yuxin Wu, Alexander Kirillov, Francisco Massa, Wan-Yen Lo, and Ross Girshick.
\newblock Detectron2, 2019.

\bibitem{xiang2016objectnet3d}
Yu~Xiang, Wonhui Kim, Wei Chen, Jingwei Ji, Christopher Choy, Hao Su, Roozbeh Mottaghi, Leonidas Guibas, and Silvio Savarese.
\newblock Objectnet3d: A large scale database for 3d object recognition.
\newblock In {\em Computer Vision--ECCV 2016: 14th European Conference, Amsterdam, The Netherlands, October 11-14, 2016, Proceedings, Part VIII 14}, pages 160--176. Springer, 2016.

\bibitem{xu2018foreign}
Haoyu Xu, Zhenqi Han, Songlin Feng, Han Zhou, and Yuchun Fang.
\newblock Foreign object debris material recognition based on convolutional neural networks.
\newblock {\em EURASIP Journal on Image and Video Processing}, 2018:1--10, 2018.

\bibitem{xu2022ppyoloeevolvedversionyolo}
Shangliang Xu, Xinxin Wang, Wenyu Lv, Qinyao Chang, Cheng Cui, Kaipeng Deng, Guanzhong Wang, Qingqing Dang, Shengyu Wei, Yuning Du, and Baohua Lai.
\newblock Pp-yoloe: An evolved version of yolo, 2022.

\bibitem{zhang2024omni6d}
Mengchen Zhang, Tong Wu, Tai Wang, Tengfei Wang, Ziwei Liu, and Dahua Lin.
\newblock Omni6d: Large-vocabulary 3d object dataset for category-level 6d object pose estimation.
\newblock In {\em European Conference on Computer Vision}, pages 216--232. Springer, 2024.

\bibitem{zhang2023densedistinctqueryendtoend}
Shilong Zhang, Xinjiang Wang, Jiaqi Wang, Jiangmiao Pang, Chengqi Lyu, Wenwei Zhang, Ping Luo, and Kai Chen.
\newblock Dense distinct query for end-to-end object detection, 2023.

\bibitem{zhao2024detrsbeatyolosrealtime}
Yian Zhao, Wenyu Lv, Shangliang Xu, Jinman Wei, Guanzhong Wang, Qingqing Dang, Yi~Liu, and Jie Chen.
\newblock Detrs beat yolos on real-time object detection, 2024.

\bibitem{zhao2019objectdetectiondeeplearning}
Zhong-Qiu Zhao, Peng Zheng, Shou tao Xu, and Xindong Wu.
\newblock Object detection with deep learning: A review, 2019.

\bibitem{zhou2024warm}
Xingcheng Zhou, Deyu Fu, Walter Zimmer, Mingyu Liu, Venkatnarayanan Lakshminarasimhan, Leah Strand, and Alois~C Knoll.
\newblock Warm-3d: A weakly-supervised sim2real domain adaptation framework for roadside monocular 3d object detection.
\newblock {\em arXiv preprint arXiv:2407.20818}, 2024.

\bibitem{zong2023detrscollaborativehybridassignments}
Zhuofan Zong, Guanglu Song, and Yu~Liu.
\newblock Detrs with collaborative hybrid assignments training, 2023.

\end{thebibliography}
\bibliographystyle{plain}

\end{document}